\documentclass[journal]{IEEEtran}
\usepackage[T1]{fontenc}
\usepackage[pdftex]{graphicx}
\usepackage{cite}
\usepackage{amsmath}
\usepackage{algorithmic}
\usepackage{array}
 \usepackage[caption=false,font=footnotesize]{subfig}
\usepackage{dblfloatfix}
\usepackage{xcolor,cite,etoolbox}
\makeatletter 
\pretocmd\@bibitem{\color{black}\csname keycolor#1\endcsname}{}{\fail}
\newcommand\citecolor[1]{\@namedef{keycolor#1}{\color{blue}}}
\makeatother
\ifCLASSINFOpdf
\else
\fi

\begin{document}
%
\title{Multi-patch Feature Pyramid Network for Weakly Supervised Object Detection in Optical Remote Sensing Images}
%
%
%
\author{Pourya~Shamsolmoali,~\IEEEmembership{Member,~IEEE,}
        Jocelyn~Chanussot,~\IEEEmembership{Fellow,~IEEE,}
        Masoumeh~Zareapoor, 
        Huiyu~Zhou,
        and~Jie Yang
\thanks{This work is supported by NSFC (No: 61876107, U1803261) and Committee of Science and Technology, Shanghai (No. 19510711200).}
\thanks{P. Shamsolmoali, M. Zareapoor,  and J. Yang are with the Institute of Image Processing and Pattern Recognition, Shanghai Jiao Tong University, Shanghai, China. (Emails: (pshams, mzarea, jieyang)@sjtu.edu.cn).}
\thanks{J. Chanussot is with LJK, CNRS, Inria, Grenoble INP, Universit\'e Grenoble Alpes, 38000 Grenoble, France. (Email: jocelyn.chanussot@grenoble-inp.fr).}
\thanks{H. Zhou is with the School of Informatics, University of Leicester, Leicester LE1 7RH, UK. (Email: hz143@leicester.ac.uk).}}

\markboth{IEEE TRANSACTIONS ON GEOSCIENCE AND REMOTE SENSING}
{Shams \MakeLowercase{\textit{et al.}}: Rotation Equivariant Feature Pyramid Network for Target Detection in Remote Sensing Imagery}
\maketitle

%
%

\begin{abstract}
\textcolor {blue}{To read the paper please go to IEEE Transactions on Geoscience and Remote Sensing on IEEE Xplore.} Object detection is a challenging task in remote sensing because objects only occupy a few pixels in the images, and the models are required to simultaneously learn object locations and detection. Even though the established approaches well perform for the objects of regular sizes, they achieve weak performance when analyzing small ones or getting stuck in the local minima (e.g. false object parts). Two possible issues stand in their way. First, the existing methods struggle to perform stably on the detection of small objects because of the complicated background. Second, most of the standard methods used hand-crafted features, and do not work well on the detection of objects parts of which are missing. We here address the above issues and propose a new architecture with a multiple patch feature pyramid network (MPFP-Net). Different from the current models that during training only pursue the most discriminative patches, in MPFP-Net the patches are divided into class-affiliated subsets, in which the patches are related and based on the primary loss function, a sequence of smooth loss functions are determined for the subsets to improve the model for collecting small object parts. To enhance the feature representation for patch selection, we introduce an effective method to regularize the residual values and make the fusion transition layers strictly norm-preserving. The network contains bottom-up and crosswise connections to fuse the features of different scales to achieve better accuracy, compared to several state-of-the-art object detection models. Also, the developed architecture is more efficient than the baselines.
\end{abstract}

\IEEEpeerreviewmaketitle

\section{Introduction}

\IEEEPARstart{H}{igh}-resolution remote sensing images (RSIs) are now available to facilitate a wide variety of applications, such as traffic management \cite{kalantar2017multiple} and environment monitoring \cite{marin2014building}. Recently, the application of object detection in RSIs has been extended from rural development to other areas, such as urban inspection \cite{cheng2016survey}, and the RSIs intensely increase in both quantity and quality. In \cite{cheng2016survey}, the authors presented a comprehensive survey and compared the performance of different machine learning methods for remote sensing image interpretation. Machine learning and deep learning-based models have been widely used for RSI object detection and classification \cite{marin2014building, hong2020graph, hong2018augmented, shamsolmoali2019novel, shamsolmoali2019convolutional }, however, supervised learning models require a large scale of annotated datasets to support satisfactory detection. Furthermore, RSI annotation needs to be undertaken by trained professionals. In contrast, weakly supervised learning is another technique to augment datasets for object detection \cite{triguero2015self, zhao2019towards}. One of the popular weakly supervised object detection methods is the combination of multiple instance learning (MIL) and deep neural networks \cite{shamsolmoali2020amil}. Despite that, their method is related to object parts rather than full body, which is due to the non-convexity of the loss functions. \par
In practice, standard machine learning models have two main stages for object detection: first feature extraction and then classification. In \cite{zhang2013object}, the extracted features were used as input, followed by support vector machine to classify the predicted targets. Standard machine learning approaches are influenced by the quality of the handcrafted and light learning-based features. Despite their promising results, standard machine learning systems fail to provide robust outcomes in challenging circumstances, for example, changes in the visual appearance of objects and complex background clutters \cite{cheng2016survey}.

In recent years, object detection is progressing due to the advance of convolution neural networks (CNNs). CNNs are able to learn features' representation using a large size of data \cite{sharif2014cnn}. An enhanced CNN introduces a process called selective search \cite{uijlings2013selective} by adapting segmentation as a selective search strategy for improving detection accuracy with higher speeds. In comparison with the analysis of natural images, the main difficulty in RSI object detection is the sizes of objects. For small objects, because of low resolution, extracting significant and discriminative features is difficult, therefore, recent works are driven towards exploring solutions of extracting discriminative features \cite{hong2020invariant}. For example, a rotational R-CNN was proposed by Guo et al. \cite{guo2020rotational} for supervised object detection. In addition, learning feature representations is a major problem in image processing tasks, and detecting multi-scale objects is challenging. To overcome this problem, pyramidal feature representations have been introduced to represent an image through multi-scale features in object detectors \cite{lin2017feature,liu2018path,ghiasi2019fpn}.
Feature Pyramid Network (FPN) \cite{lin2017feature} is among the best representative approaches for producing pyramidal feature representations of objects. Typically, pyramid models adopt a backbone network that is built for image segmentation or classification and to create feature pyramids by successively merging two or three consecutive layers in the backbone network with top-down and adjacent connections. High-level features have lower resolutions but they are semantically strong and can be upscaled and merged with higher resolution features to create more discriminative features.
Scholars who are working on object detection in RSIs have observed the powerful ability of FPN, and applied this method to their works. Li et al. \cite{li2018hsf} proposed a feature-based method for identifying ships in RSIs. The authors proposed a region based network to detect ships from the generated feature maps. In spite of its object detection performance, its computation is inefficient. In \cite{li2019nested}, a deep network based upon two-stream pyramid module was proposed for detecting multi-scale salient objects in RSIs. Their network has demonstrated promising performance on detecting objects of a regular size but failed to perform accurately on small-size or incomplete objects. In \cite{yang2019cdnet}, the authors proposed a detector based on an encoder-decoder FPN for detecting clouds from RSIs. The network is simple and efficient, but not effective to the crowd environments.
Although FPN is effective and simple, it may not be an efficient architectural design. Recently, in \cite{ghiasi2019fpn, li2018hsf}, the authors added extra bottom-up and skip connection pathways onto FPN to enhance feature representations. Nonetheless, these methods only take into account one of the three dimensions (image size, depth, and width). However, by analyzing more dimensions and adopting fusion methods \cite{hong2020more, yokoya2017hyperspectral}, we can train a network to achieve better performance and efficiency. Shamsolmoali et al. \cite{shamsolmoali2008road} recently introduced a new architecture named SPN that improves the performance of FPN by extracting effective features from all the layers of the network for describing different scales’ objects. This architecture is introduced to learn from the multi-level feature maps and improve the semantics of the features.

To tackle the above problems, in this paper, we introduce a scalable architecture to build effective pyramidal representations, named MPFP-Net. More specifically, the proposed model mainly consists of two components. First, we proposed a multiple patch learning (MPL) scheme to deal with the non-convexity and lack of full object representation. MPL treats input images as patches. During training, MPL learns patch subsets, which have mutual co-relation. Patch subsets with accurate trailer parameters can activate semantic regions to describe a full object. Second, we propose a new pyramid network with cross-scale connections for producing multi-scale representative features. An additional advantage of the modular pyramidal architecture is the capability of efficient multi-scale object detection. There are three contributions made in this paper.
\begin{itemize}
\item{Our proposed MPL strategy is built on a CNN to accurately describe an object such that the network performs even robustly with different backbone models, such as ResNet \cite{he2016deep} and SPN \cite{shamsolmoali2008road}. By adopting SPN as the backbone into the proposed model, the detection accuracy and speed of MPFP-Net are better than those of the other state-of-the-art models.}
\item{We incorporate scale-wise image fusion within the FPN architecture. The cross-scale connections are used to adapt the number of the channels and the size of the feature maps to maintain the norm of the gradients. Also, we propose a computationally efficient method to extract and normalize a multi-level feature response to the corresponding level. By using a layer for extracting multi-scale feature maps, the network weights are updated once per iteration, which considerably improves the training efficiency of the proposed model.}
\item{We fully evaluate several recent deep CNNs for object detection in RSIs and the overall performance is reported.}
\end{itemize}
The rest of this paper is structured as follows, Section~\ref{sec:2} shows a brief review of the existing object detection methods on RSIs. In Section~\ref{sec:3}, we describe our proposed architecture in detail. In Section~\ref{sec:4}, we report the experimental results on the RSI datasets. Section~\ref{sec:5} contains the ablation study and we conclude the paper in Section~\ref{sec:6}. 
\section{RELATED WORK}
\label{sec:2}
In the last ten years, object detection in RSIs has achieved significant progress. In this section, we will discuss the available object detection technologies for natural and RS images. In particular, we focus on the discussion on weakly supervised models for object detection.
\subsection{Object Detection in Nature Scene Image}
Deep CNNs significantly improves the performance of object detection models in recent years. In \cite{girshick2015region}, the authors proposed a detector named R-CNN. R-CNN has a high detection rate, but its speed is limited. With the intention of increasing accuracy and speed of object detection, Fast R-CNN \cite{girshick2015fast} was proposed to use bounding-box regression with end-to-end training. Later, Faster R-CNN \cite{ren2015faster} was introduced to combine object proposals and identification within a unified network with impressive efficiency. Generally, there are two approaches to deal with the scale-variation problem. The first approach is to use feature-wise image pyramids to generate semantical multi-scalar features. Features from the images of different scales return individual predictions to generate the final detection. With regard to localization precision and detection accuracy, features from the images of different sizes outweigh the features that are only based on the images of a single size \cite{guo2020rotational}. 
\begin{figure*}
  \centering
  \includegraphics[width=6.1in]{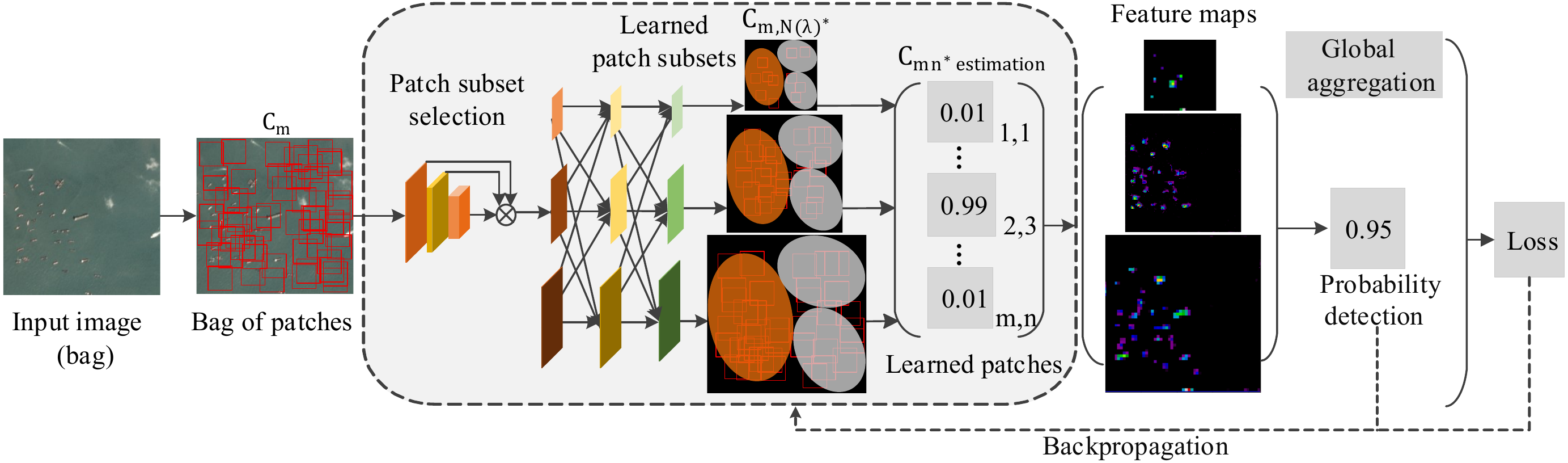}
\caption{The overall architecture of the proposed Multiple Patch Feature Pyramid Network, it contains patch detection, subsets selection, feature extraction and instance classification.}
\label{fig:2}
\end{figure*}
The second approach is to detect objects in the feature pyramid (FP) extracted from different layers in the network while as an input only a single-scale image is applied. This methodology requires a smaller memory space than the first one. Additionally, the standard FP unit can be modified and inserted into the state-of-the-art CNNs based detectors. Lin et al. \cite{lin2017feature} proposed a feature pyramid network with double-path connections to seek more representative features. Ding et al. \cite{ding2020weakly}  proposed a weakly supervised pyramid CNN for object detection. The model consists of a hierarchical configuration with both top-down and bottom-up connections to learn high-level semantics as well as low-level features. 
\subsection{Object Detection in RSI}
Compared to object detection in natural scene images, object detection in RSIs has additional challenges. This topic has been intensively studied over the last few years. Conventional object detection models have ranking systems to categorize objects and background \cite{qiu2017automatic}. For example, in \cite{wu2019fourier,wu2019orsim}, the authors proposed an efficient detection framework based on rotation-invariant feature aggregation and adopted a learning method to acquire significant and meaningful features for small object detection.

With deep learning development, there has been significant improvement on object detection. In \cite{cheng2016learning}, the authors proposed an equivariant CNN method by introducing a regularization method for object detection in RSIs. Dong et al. \cite{dong2019sig} proposed a deep transfer learning method on the basis of R-CNN to minimize the loss of tiny objects in RSIs. Moreover, transfer learning can be used to label RSIs by annotating both object positions and classes. In \cite{deng2018multi}, the authors proposed multi-scale object detection models in RSIs using inspection modules to train the sub-networks. The proposed model has significant detection performance on multi-scale objects but is not efficient due to the depth of the network.
Different conditions and objects variations in RSIs bring various challenges to object detection in such images. Consequently, it is hard to obtain satisfactory results by deploying the available object detection models. Furthermore, due to the deficit of training data and the complexity of the network architecture to handle various objects with multiple scales and complex background, we introduce a novel multi-scale deep learning model for object detection in RSIs.
\subsection{Weakly Supervised Object Detection}
Multiple instance learning (MIL) is a weakly supervising learning method that, in the training phase, treats each image as a "bag" and constantly picks high scored instances from the bags. It performs similarly to an Expectation-Maximization algorithm for estimating instances and training detectors concurrently. On the other hand, such a model, due to lack of local minima in the non-convex loss functions, does not have smooth training process \cite{wan2018min}. To mitigate the non-convexity issue, bundling was used as one of the pre-processing approaches to simplify the selection of relevant instances \cite{song2014learning}. A box-level supervision method was presented to decrease the solution space throughout a recurrent constraint network \cite{zhao2019towards}. In \cite{bilen2016weakly}, MIL-Net was proposed in which the convolution operator and filters are used as detectors. Nevertheless, MIL-Net's loss functions stay non-convex. To deal with this problem, spatial regularization is used \cite{bilen2016weakly} within the cascaded convolutional networks. Current approaches mainly use selective regions (patches) as ground-truth to gradually improve the classifiers \cite{tang2017multiple}. Existing methods that use spatial regularization and gradual refinement are successful at enhancing object detection. Even so, it is lack of a systematic approach to deal with the local minimum problem.

\section{PROPOSED MODEL AND METHODOLOGY}
\label{sec:3}
In this section, we present the details of our proposed model that consists of two main components: global multiple patch (image regions) learning and a variation of multi-scale feature pyramid networks with a novel fusion scheme to robustly detect an object region to further improve the detection performance. Fig. \ref{fig:2} shows the architecture of our proposed model.
\vspace{-1mm}
\subsection{Multiple Patch Global Learning (MPL)}
\label{sub:1}
In MPL, images are treated as bags and the image's regions as patches. In MPL, the labels are only allocated to bags of patches. Here, $C_m\in C$ shows the $m^{th}$ bag and $C$ represents all the bags (training images). $y_m\in Y$ while $Y$=$[1, -1]$ shows the bag's label, and $C_m$ shows the possibility that the bag owns positive patches. $y_m=1$ signifies a positive bag which always holds a single patch. On the other hand, $y_m$= $-1$ signifies a negative bag while the entire patches are negative. Let $C_{mn}$ and $y_{mn}$ indicate the patches and labels in bag $C_m$ , in which $n\in \{1, 2,..., N\}$, $N$ is the number of the patches.
The MIL methods have two steps for object detection: patch selection (patch controller) as the initial step and object estimation \cite{wan2019c, zhao2019towards}. In the patch selection phase, a patch controller $f (C_{mn}, \omega_f)$ calculates the object score of all the patches to obtain a patch from $C_m$. 
\begin{eqnarray}
C_{mn^\ast}=argmax_nf(C_{mn}, \omega_f),
\end{eqnarray}
in which $\omega_f$ denotes the parameters of the patch selector and $n^\ast$ denotes the highest scored patch. With certain patches, a detector $d_z (C_{mn}, \omega_d)$ with parameter $\omega_d$ is trained, in which $z\in Y$. $\omega_d$ represents the parameters of the object detector. In standard MIL methods \cite{zhao2019towards, bilen2016weakly}, the above processes are combined and $f (C_{mn}, \omega_f)$ and $d (C_{mn}, \omega_d)$ are fully integrated as follows:
$\setlength{\arraycolsep}{0.0em}$ 
\begin{eqnarray}
Loss (C,\omega)=\sum_m Loss_f (C_m, \omega_f)+  \nonumber\\ 
Loss_d (C_m, C_{mn}, \omega_d),
\label{eq:2}
\end{eqnarray}
$\setlength{\arraycolsep}{0.0em}$
in which the standard loss of patch selection is determined as
\begin{eqnarray}
Loss_f (C_m, \omega_f)=max(0, 1-y_m max_n f(C_{mn}, \omega_f)), 
\label{eq:3}
\end{eqnarray}
and the loss of detector estimation is determined as
$\setlength{\arraycolsep}{0.0em}$ 
\begin{eqnarray}
Loss_d (C_m, C_{mn^\ast}, \omega_d)=\nonumber\\
-\sum_z \sum_j \delta_{z,y_{mn}}\log d_z (C_{mn}, \omega_d), 
\label{eq:4}
\end{eqnarray}
in which $y_{mn}$ is determined based on the metric introduced in \cite{everingham2010pascal}:
\begin{eqnarray}
y_{mn}=
\begin{cases}
+1 & \text{if  IoU ($C_{mn}, C_{mn^\ast}$) $\geq 0.5$}\\
 -1 & \text {if  IoU ($C_{mn}, C_{mn^\ast}$) $< 0.5$}
\end{cases}
\label{eq:5}
\end{eqnarray}
for $m = n$ or $m\neq n$ the Delta function $\delta_{mn}= 1$ or $0$ respectively.
\begin{eqnarray}
\delta(mn)=\lim_{\sigma \rightarrow 0}\frac{1}{\sigma \sqrt 2\pi}
\label{eq:6}
\end{eqnarray}
To enhance the performance of the patch selector, we introduce a novel optimization technique. This approach splits the patches into subsets whilst handling the non-convexity of the loss function shown in Eq.(\ref{eq:2}).
Our model, includes a sequence of smooth loss functions from the beginning point ($\omega^0, 0$) to the resulting point ($\omega^\ast, 1$), where $\omega^0$ is the result of $Loss (C, \omega, \Gamma)$ where $\Gamma=0$, and $\omega^\ast$ is used where $\Gamma = 1$. Thus, we determine a range of $\Gamma$, $0 = \Gamma0 <\Gamma1 <...<\Gamma \tau = 1$, and accordingly improve Eq.(\ref{eq:2}) to a persistent loss function, as follows:
$\setlength{\arraycolsep}{0.0em}$
\begin{eqnarray}
\omega^\ast = \underset{\omega_f, \omega_d}{\arg\min}\sum_m Loss_f (C_m, C_{m,N(\Gamma)}, \omega_f) \nonumber\\
+Loss_d (C_m, C_{m, N(\Gamma)}, \omega_d)
\label{eq:7}
\end{eqnarray}
in which $C_{m, N(\Gamma)}$ represents the patch subset and $N(\Gamma)$ the index of $C_{m, N(\Gamma)}$ , controlled by $\Gamma$ which is a continuous parameter. $Loss_f (C_m, C_{m, N(\Gamma)}, \omega_f)$  is the persistent loss function of patch selection, and $Loss_d (C_m, C_{m, N(\Gamma)}, \omega_d)$  is the persistent loss function of the detector.
To learn the patch selector, a bag (image) is divided into patch subclasses. In each subclass, the objects are spatially related and may have overlapping with each other, and each class contains partially similar objects. Each subclass at least contains a single bag (image) $C_{m,N}\cup C_{m,\acute N}=C_m$, and $C_{m,N}\cap C_{m,\acute N} = \emptyset$ for $\forall \, N\neq \acute N$. 
The overall patches in an image (bag) are arranged by their object scores $f (C_{m,n}, \omega_f)$ and the following two processes are carried out: 1) Create a patch for a  subclass based on the highest object score. 2) Identify the patches that are overlapped with the highest scored patch $C_{m,n^\ast}$ and then include them in the subclasses. The successive patch selection is made with $0\leq \lambda \leq 1$ in which the loss function is formulated as follows:
\vspace{-2mm}
$\setlength{\arraycolsep}{0.0em}$
\begin{eqnarray}
Loss_f (C_m, C_{m, N(\lambda)}, \omega_f)=\nonumber\\
\max(0, 1-y_m \, \underset{N(\lambda)}{\max}f(C_{m, N(\lambda)}, \omega_f)), 
\label{eq:8}
\end{eqnarray}
where $f(C_{m, N(\lambda)}, \omega_f)$, the score of patch subclass $C_{m, N(\lambda)}$, is determined as follows:
\begin{eqnarray}
f(C_{m, N(\lambda)}, \omega_f)= \frac{1}{\vert C_{m, N(\lambda)}\vert} \sum_n f(C_{mn}, \omega_f), 
\label{eq:9}
\end{eqnarray}
in which $\vert C_{m, N(\lambda)}\vert$ indicates the total number of the patches in subclass $C_{m,N(\lambda)}$ and $C_{mn}\in C_{m,N(\lambda)}$.
In the time of model learning, all the patches in subclass $C_{m, N(\Gamma)}$ equally engage to adjusting the network parameters. When $\lambda=0$, each bag $C_m$ only has one subset that contains all the patches. Given $\lambda=1$, $C_m$ is divided into several subsets, where each subset bears at least a single patch and consequently Eq.(\ref{eq:8}) is not satisfied. Referring to Eq.(\ref{eq:9}), the score of a patch in general is similar to the mean score of the patches in that subset. Therefore, our defined loss function Eq.(\ref{eq:7}), is convex and grows linearly, compared to the standard MIL Eq.(\ref{eq:3}).
In supervised learning, for the subclass $C_{m,N(\lambda)}$, the maximum average score is taken for object detection. If bounding-box annotation is not feasible, the patch selector may not be accurate and the selected subclass may contain the background or only part of the objects. To fully outline valid patches and learn to detect objects, the patches are divided into positive and negative categories with parameter $\lambda$. Let $C_{m, N(\lambda)^\ast}$ be the learned patch subclass and the patch of the highest score in $C_{m, N(\lambda)^\ast}$ be $C_{mn^\ast}$. Patches are separated into positive and negative categories on the basis of their relations, as follows:
\begin{figure*}
  \centering
  \includegraphics[width=0.82\textwidth]{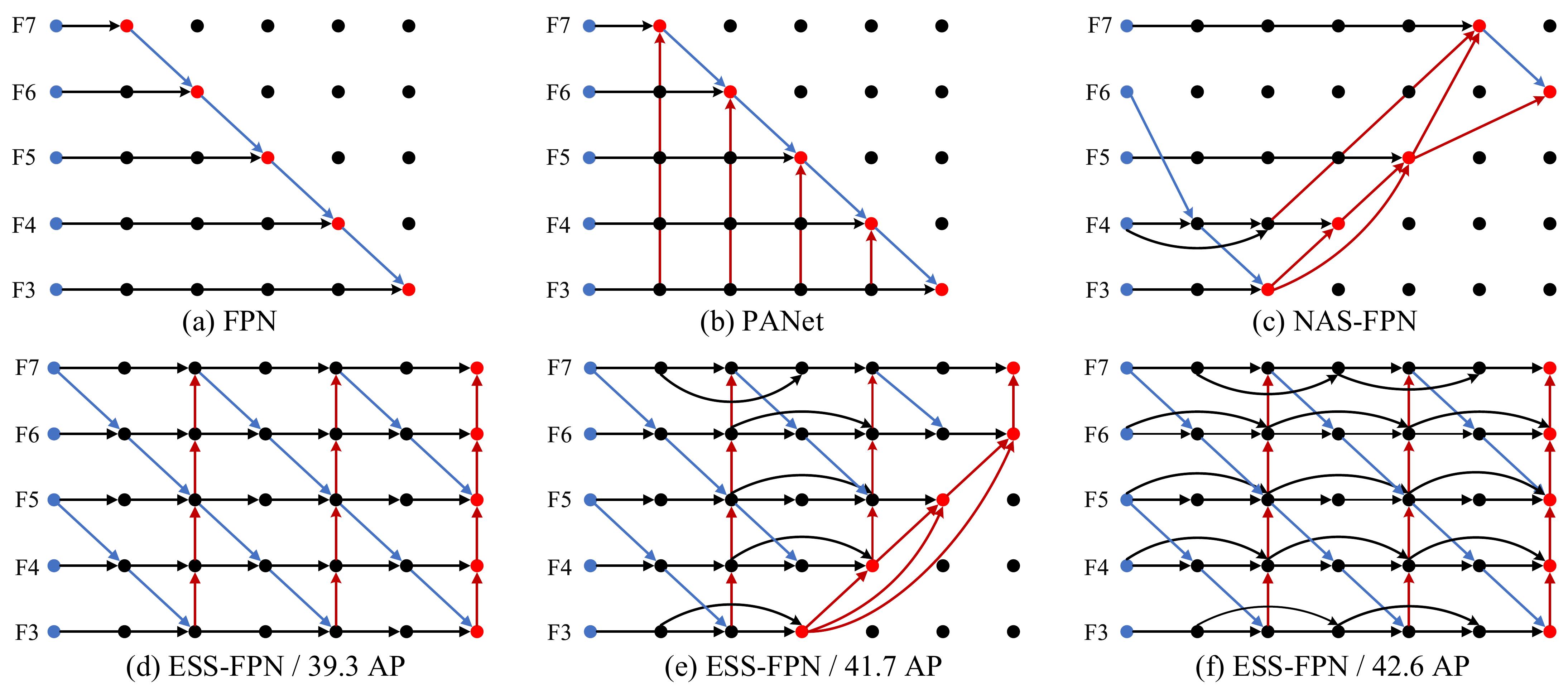}
\caption{Proposed network design where the top-down paths are shown in {\textcolor{blue}{blue}} and bottom-up paths are shown in {\textcolor{red}{red}} – (a) FPN \cite{lin2017feature} shows a top-down path for fusing multi-scale features (F3 - F7); (b) PANet \cite{liu2018path} added additional bottom-up paths to the FPN; (c) NAS-FPN \cite{ghiasi2019fpn} adopted neural architecture search to identify a feature network; In (d), (e) and (f), we shown different architectures of ESS-FPN learned by multi-scale feature fusion. (f) The architecture of ESS-FPN used in the experiments.}
\label{fig:4}
\end{figure*}
\begin{eqnarray}
y_{mn}=
\begin{cases}
+1 & \text{if  IoU ($C_{mn}, C_{mn^\ast}$) $\geq 1- \lambda/2$}\\
 -1 & \text {if  IoU ($C_{mn}, C_{mn^\ast}$) $< \lambda/2$}
\end{cases}
\label{eq:10}
\end{eqnarray}
in which $IoU$ stands for the intersection of union between two consequent patches. Eq.(\ref{eq:10}) states that patches where their IoU are above the threshold, ($1-\lambda/2$) are positive and the patches with $C_{mn^\ast}$ less than $\lambda/2$ are negative. As stated in Eq.(\ref{eq:10}), gradually the other patches are identified as positive or negative and the detector $d_z (C_{mn}, \omega_d)$  in accordance with these patches, predicts the objects with the following loss function:
$\setlength{\arraycolsep}{0.0em}$
\begin{eqnarray}
Loss_d (C_m, C_{m, N(\lambda)},\omega_d )= \nonumber\\
\sum_z\sum_n \delta_{z,y_{mn}} \log d_z(C_{mn}, \omega_d). 
\label{eq:11}
\end{eqnarray}
\vspace{-5mm}
%
%
\subsection{Feature Pyramid Network for Patch Estimation}
\label{sub:2}
Before applying classification to the whole image, we aim to generate image regions (patches) by randomly extracting patches from the images. In the next step, the collected patches go through the proposed FPN to estimate the probability of the objects. Consider an image $X$ with a dimension $H\times W$ pixels. If the sliding window has $K\times K$ pixels and $d$ denotes the stride, then $X$ is divided to create the patches $\{c_{i, j}\}$, in which $i\in \{1, 2, 3,..., \lceil(H-K)/d\rceil+1\}$ and $j\in \{1, 2, 3,...,\lceil(W-K)/d\rceil+1\}$ represent the vertical- and horizontal-axis respectively. These indexes report the information of the patches. Each bag (image) contains different numbers of patches and the patches can be resized to match the network's input size.
For the patch-level estimation, we propose a novel FPN model that contains two main components: a backbone network for constructive feature extraction, and a scale-wise feature fusion module to efficiently integrate high level semantic features with low-level ones. The proposed model aims to optimize feature fusion with a better inherent way to improve object detection in RSIs. In the next section, we target at the problem of multi-scale feature fusion, and propose a novel model with multiple cross-scale connections and weighted feature fusion. Our efficient scale-wise feature pyramid architecture is named ESS-FPN.

\subsubsection{Multi-scale Feature Fusion}
The goal of multi-scale feature fusion is to combine different scales of features. $\vec F^{in}=(F_{l_1}^{in}, F_{l_2}^{in},...)$ represent the multi-scale feature list, in which $F_{l_i}^{in}$ indicates the $l_i^{th}$ feature. Our objective is to transform $f$ to effectively combine multi-scale features and generate a list of features as the output: $\vec F^{out}=f(\vec F^{in})$. Fig. \ref{fig:4}(a) represents the methodology of the standard top-down FPN \cite{lin2017feature}. This model gets the input features of level (3-7) $\vec F^{in}=(F_3^{in},...,F_7^{in})$, while $F_i^{in}$ signifies a feature with $1/2^i$ resolution of the input images. As an example, if the input resolution is of $1024\times1024$, then $F_3^{in}$ indicates the third level feature $(1024/2^3=128)$ with the resolution of $128\times128$, where $F_7^{in}$ shows the seventh level feature with the resolution of $8\times8$. The standard FPN combines the features in a hierarchical fashion: 
\begin{equation} 
\left\{\,
\begin{IEEEeqnarraybox}[][c]{l?s}
\IEEEstrut
F_7^{out}=conv(F_7^{in}) \\
F_6^{out}=conv(F_6^{in}+ rescale(F_7^{out}))  \\
... \\             
F_3^{out}=conv(F_3^{in}+ rescale(F_4^{out}))   
\IEEEstrut
\end{IEEEeqnarraybox}
\right.
\label{eq:example_left_right1}
\end{equation}
in which rescaling is often a down/up scaling operation to match the image resolution, and Conv denotes a convolutional operation for the feature processing.

\subsubsection{Cross-Wise Feature Fusion}
The standard top-down FPN cannot widely reuse the features that are extracted in the previous layers as it only contains single-path information flows. Several approaches have been proposed for handling this problem. In PANet \cite{liu2018path}, the authors proposed to use both the top-down and reversed paths in the network architecture, as presented in Fig. \ref{fig:4}(b). NAS-FPN \cite{ghiasi2019fpn} introduced a FPN, based on neural network search, to develop a scale wise network architecture, however it is not efficient to search for the best path and it is challenging to modify the architecture, as indicated in Fig. \ref{fig:4}(c). After having evaluated the accuracy and efficiency of these three networks, NAS-FPN achieves higher accuracy than FPN and PANet but with expensive computation. To enhance the performance and efficiency of our model, the nodes that contain single input edges are eliminated. If a particular node has a single input edge, then its contribution to the feature network is subtle and can be used in a fusion scheme. This has resulted in a multi-path network. 

Unlike PANet and NAS-FPN that only have top-down and bottom-up paths, in our bottom up and cross-vise model, each path is treated as a single feature network layer. The same layer repeats several times in various directions and at the final stage the features at different scales are fused to generate high-level feature fusion. 
To fuse a number of input features with various resolutions, a simple yet effective method re-configures them to be of equal resolutions and then merges them. PAN \cite{liu2018path} proposed global upscaling for improving pixel localization. Standard feature fusion methods equally treat the overall input features. Nevertheless, as the features of the input images have various resolutions, they do not share equal contributions to the output feature. To deal with this problem, we propose that each input has an additional weight during the feature fusion. In light of this idea, the following three weighted fusion approaches are considered: \par

{\bf Weighted Feature Fusion:} as demonstrated in \cite{hu2019dynamic}, a learnable weigh can improve the accuracy with very small computation costs. Nevertheless, due to the unbounded scalar weight, it may lead to unstable training. For this reason, weight normalization is generally used to limit the range of values for weights. As discussed earlier, the outputs of the top-layers are much similar to the ground-truth. Using the same weights for all the locations compels the network to learn better fusion weights, which unfortunately miss the contributions from the low-level features. This prevents low-level features delivering adequate edge information, which is useful to identify object boundaries. Thus, before feature fusion, we deal with the scale variation of different level responses via normalization of their scales. Thereby, a robust weight learner can avoid scale variation and better learn fusion weights. A systematic multi-level feature extractor with a normalizer is here proposed to extract and normalize multi-level responses to the same scale (see Fig. \ref{fig:4}(e) and \ref{fig:4}(f)). In particular, feature normalization unit in the module is in charge for feature map normalization in each layer. In the proposed dynamic fusion model, two different schemes for predicting adaptive fusion weights are devised to determine the location fusion weights and ensure location-invariance. The model equally treats all the feature maps and global fusion weights are learned subsequently. Next, the model adjusts multi-level cross outputs A$_{cross}$ of size $H\times W$, to acquire a fused output $A_{fuse}=f(A_{cross})$ by combining multi-level and multi-scale outcomes. The fusion process is as follows:
\begin{figure*}
  \centering
  \includegraphics[width=6.7in]{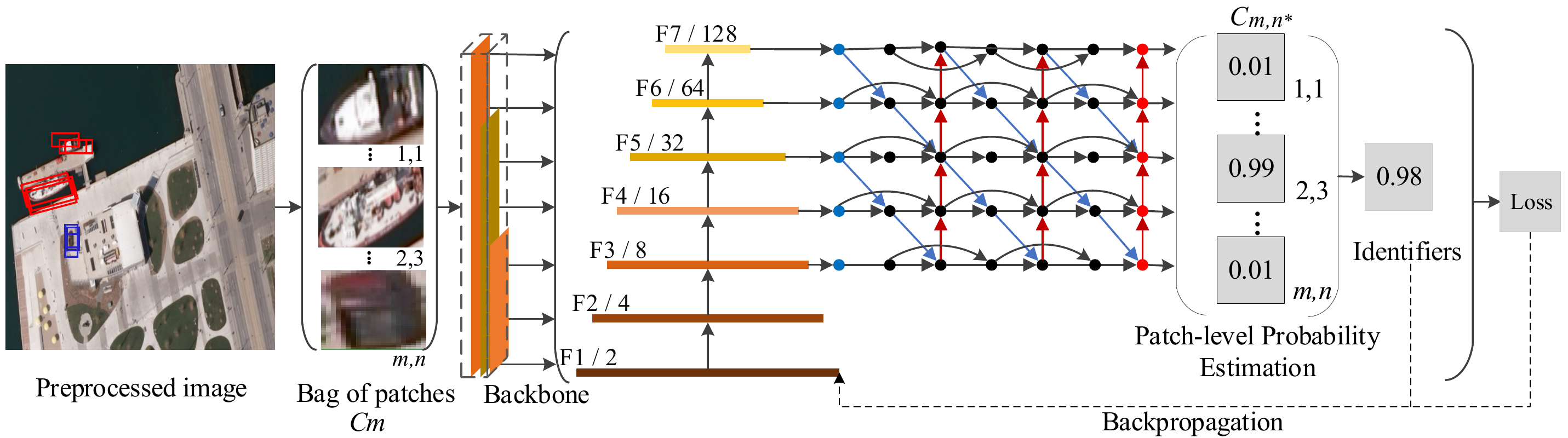}
\caption{The proposed network contains: patch detection, it adopts ResNet-50 + SPN \cite{shamsolmoali2008road} as the backbone, ESS-FPN as the multiscale feature network, and class prediction network.}
\label{fig:5}
\end{figure*}
$\setlength{\arraycolsep}{0.0em}$
\begin{eqnarray}
A_{fuse}^i=w_1^i A_{cross6}^i + w_2^i A_{cross5}^i + \nonumber\\
w_3^i A_{cross4}^i +  w_4^i A_{cross3}^i 
\label{eq:12}
\end{eqnarray}
in which $(w_1^i,w_2^i,...,w_5^i)$  are the parameters of the $k^{th}$ $1\times1$ convolution layer, denoting the fusion weights for the $i^{th}$ levels. The above equation can be generalized to the following form:
\vspace{-2mm}
\begin{eqnarray}
A_{fuse}=f(A_{cross}; W)
\label{eq:13}
\end{eqnarray}
in which $f$ summarizes the operation of Eq.(\ref{eq:12}) and $W=(w_1^i,w_2^i,...,w_5^i)$ represents the fusion weights. In addition, an adaptive weight learner is proposed to learn the fusion weights based on the feature condition as follows:
\begin{eqnarray}
A_{fuse}=f(A_{cross}; \Phi(x)),
\label{eq:14}
\end{eqnarray}
where $x$ indicates the feature map. The above equations illustrate the key differences between our designed adaptive weight fusion model and the current static weight fusion model. The fusion weight $W$ is strongly based on the feature map $x$, i.e. $W=\Phi(x)$. Each input feature map $x$ will engender different parameters $\Phi(x)$ and consequently adjust this adaptive weight learner. Thus, the model can quickly adjust to the input image and satisfactorily learn multi-level outcome fusion weights in an end-to-end mode. 
$\Phi(x)$ represents two fusion weight strategies, location-invariant weight learners and location-adaptive weight. The location-invariant weight learner totally learns $5k$ fusion weights that are generated from each location of the fused feature maps:
\begin{eqnarray}
\Phi(x)=(w_1^i,w_2^i,w_3^i,w_4^i, w_5^i), i\in [1, k].
\label{eq:15}
\end{eqnarray}
On the other hand, the location-adaptive weight learner generates $5k$ fusion weights in accordance with different spatial locations, which in sum leads to $5k\times H\times W$ weighting parameters.
$\setlength{\arraycolsep}{0.0em}$
\begin{eqnarray}
\Phi(x)=(W_{s, t}),  \nonumber\\
W_{s,t}=((w_1^i)_{s, t},(w_2^i)_{s, t}, (w_3^i)_{s, t},(w_4^i)_{s, t}, (w_5^i)_{s, t}).  
\label{eq:16}
\end{eqnarray}
where $s\in [1, H]$, $t\in [1, W]$ and $i\in [1, k]$. In this model, the location-invariant weight learner generates global fusion weights, whereas the location-adaptive weight learner aggregates the generated fusion weights for each location based on the spatial variations. \par

{\bf Softmax fusion:} If multiclass objects exist, we apply softmax to all the weights, therefore all the weights are normalized to the range between 0 and 1, indicating the importance of each input. Then, the weighted feature maps are aggregated across all the scales to build the fused feature maps. Multi-scale features $f_{l,c}^s$ in which $l, c, s$ are the indexes of locations, channels and scales respectively. The attention model $F$ is notated as $F(f_l^1,...,f_l^S)$. The fused feature map is formulated as:
\begin{eqnarray}
\hat f_(l,c)=\sum_{s=1}^S \hat w_{l,c}^s . f_{l,c}^s 
\label{eq:17}
\end{eqnarray}
Among all the previous works \cite{wan2019c} where sigmoid is used for selecting features from different scales $w_l^s=\frac{exp(h_l^s)}{\sum_{t=1}^s exp(h_l^t)}$, in our model, softmax weight is used as $\hat w_{l,c}^s=\frac{1}{1+e^{-h_{l,c}^s}}$. Since the proposed feature fusion model is performed across all the layers instead of only the final layer \cite{liu2018path}, we realised that softmax has better performance in fusing multi-scale features across different layers. However, having softmax causes additional costs. To reduce the extra costs, we introduce an instant fusion approach.\par

{\bf Instant fusion:} In this approach, softmax is not used, therefore, the process is faster. In this operation, the total value of each weight falls between $0$ and $1$. The instant fusion is formulated as:
\begin{eqnarray}
I_f=\sum_i \frac{w_i}{\theta + \sum_j w_j}. I_i, 
\label{eq:18}
\end{eqnarray}
In which $w_i\geq0$ is obtained using the standard Relu activation function, and to prevent numerical instability, a small value  $\theta=0.001$ is used. Our experiments illustrate that our instant fusion approach obtains equal results as the sigmoid-based fusion, but runs up to 25\% faster. In the proposed model, we combine the proposed multi-directional connections and the instant fusion to obtain highly semantic features. Here, we demonstrate the details of the feature fusion in the $5^{th}$ layer as illustrated in Fig. \ref{fig:4}(f):
\begin{eqnarray}
F_5^{td}=Conv(\frac{w_1.F_5^{in}+w_2.rescale(F_6^{in})}{w_1+w_2+\theta}),
\label{eq:19}
\end{eqnarray}
\begin{eqnarray}
F_5^{out}=Conv(\frac{\acute w_1.F_5^{in}+\acute w_2.F_5^{td}+\acute w_3.rescale(F_4^{out})}{\acute w_1+\acute w_2+\acute w_3+\theta}).
\label{eq:20}
\end{eqnarray}
in which $F_5^{td}$  is the transitional feature at the fifth  level of the cross route, and $F_5^{out}$  is the result of the fifth layer on the upward route. All the rest of the features are constructed similarly. It is worthy to mention, multi-scale divisible convolutions are adjusted to the network to improve the efficiency.
\begin{table} 
\centering
  \caption{SCALING CONFIGURATION FOR ESS-FPN $(S_0 - S_6) - \Phi$ IS THE COEFFICIENT PARAMETER THAT ADJUSTS THE OTHER SCALING DIMENSIONS.}
  \label{tab:1}
  \begin{tabular}{lccc}     \hline
\\[-0.7em]
   & Input size& W$_{essfpn}$& D$_{essfpn}$ \\    \hline
\\[-0.7em]
$S_0(\psi = 0)$ & 512 & 32 & 3 \\
$S_1(\psi = 1)$ & 640 & 64 & 4 \\
$S_2(\psi = 2)$ & 768 & 88 & 5 \\
$S_3(\psi = 3)$ & 896 & 112 & 6 \\
$S_4(\psi = 4)$ & 1024 & 160 & 7 \\
$S_5(\psi = 5)$ & 1280 & 224 & 7 \\
$S_6(\psi = 6)$ & 1280 & 288 & 8 \\
\hline
\end{tabular}
\end{table}
\subsubsection{ESS-FPN Architecture}
Fig. \ref{fig:5} presents the architecture of ESS-FPN. SPN \cite{shamsolmoali2008road} is used as the backbone. The proposed ESS-FPN acts as the network for multi-scale feature extraction that receives the features from the $3^{rd}$ to $7^{th}$ layer of the backbone and constantly executes top-down, bottom-up and cross-wise feature fusion. To improve both accuracy and efficiency, the current approaches scale up a baseline by adopting larger backbone networks, using a larger dataset, or adding more FPN layers \cite{yang2019cdnet, he2016deep, tan2020efficientdet}. Such models are generally unproductive as they can only process one or few scaling dimensions (i.e. depth, width, and input resolution). We propose an efficient scaling method for object detection, which adopts an extensive multiplex factor $\psi$ to rescale the entire dimensions of the backbone network. Different from other image classification models \cite{real2019regularized}, object detectors have various scaling dimensions, therefore searching over all the dimensions is considerably expensive. Thus, we scale up the dimensions by using a heuristic method. 
We increase the ESS-FPN's depth (layers) $D_{essfpn}$ and the width (channels) $W_{essfpn}$, to improve the proposed detector's efficiency. More precisely, grid search is applied on the bases of $\{1.15, 1.2, 1.25, 1.3, 1.35, 1.4\}$, and we choose the premier value $1.3$ as the adjustment factor for the width. Conventionally, the width and depth of ESS-FPN are determined as follows:
\vspace{-2mm}
\begin{eqnarray}
D_{essfpn}=3+ \psi , \; \; \;  W_{essfpn} = \frac{2^7}{1.3^\psi}
\label{eq:21}
\end{eqnarray}
We use feature levels 3-7 in ESS-FPN, therefore the input resolution should be dividable by $2^6=64$, and the following equation is used:
\vspace{-2mm}
\begin{eqnarray}
Res_{input}=\frac{512+\psi}{64}
\label{eq:22}
\end{eqnarray}
where, with various $\psi$, we have performed a wide range of evaluations in order to find the best parameter from $S_0(\psi = 0)$ to $S_7(\psi = 7)$ as listed in Table \ref{tab:1}, in which $S_7$ has a higher resolution than $S_6$. 
For each processed image patch, the ultimate sigmoid layer will generate a probability distribution that will be used for further patch selection. Compared to the other FPNs \cite{lin2017feature, ghiasi2019fpn}, in the proposed MPFP-Net inspired by SPN \cite{shamsolmoali2008road}, U-shape network is adopted to build a pyramid network. In this model, the upward path gets the outputs of multiple layers as its reference sets.

 Further, to enhance the performance and preserve the features' smoothness, $1\times1$ convolution layers are inserted after each up-scaling process. In total, 5 stacks of U-shape networks are used for building the multi-scale features. In our model, after the patch-level estimation, visual domain aggregation is applied to connecting all the detected patch-level probabilities to a detected image map. In the present patch, visual features are first aggregated using a pooling layer and then transformed to a semantic domain.

\section{EXPERIMENTS}
\label{sec:4}
\begin{figure}
\vspace{-7mm}
  \centering
  \includegraphics[width=3.4in]{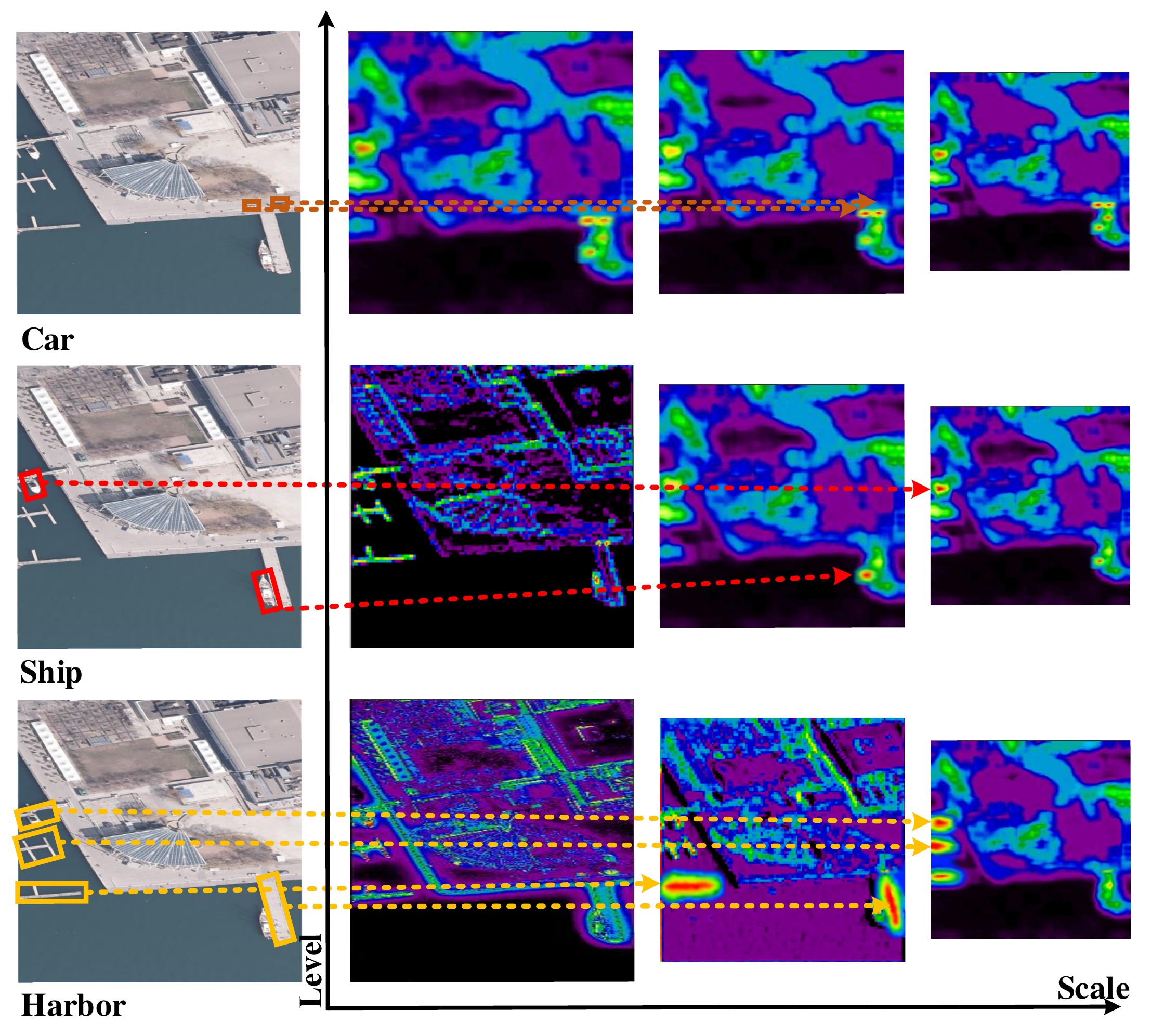}
\caption{Examplar activation values of multi-scale multilevel features. Best view in color.}
\label{fig:7}
\end{figure}

In this section, firstly we discuss the details of the benchmark datasets for object detection in RSIs and then describe the evaluation metrics, training procedure, and implementation details of our proposed model. Next, we compare the performance of the proposed MPFP-Net with that of several state-of-the-art approaches. MPFP-Net is implemented in PyTorch and all the experiments are conducted on a workstation equipped with Tesla P40 GPU.
To validate that the proposed MPFP-Net can learn effective features for object detection with different appearance variations and scales, the activation values of the classification convolution layers including scales and level dimensions are shown in Fig. \ref{fig:7}. The input image contains four harbors, two ships and two cars. The sizes of harbors, ships and cars are different. 

It is worth mentioning that: (1) compared with the smaller harbors, the larger harbors have the larger activation value at the feature map of a large scale, similar to the larger ship and the smaller one; (2) the smaller harbors and the smaller ship have higher activation values at the feature maps of the same scale. This sample illustrates that: MPFP-Net learns effective features to deal with different scales and appearance-complexity across object patches. MPFP-Net is evaluated on three public datasets: NWPU VHR-10 \cite{dong2019sig}, LEVIR \cite{zou2017random}, and DOTA \cite{xia2018dota}.
\begin{table*} 
\centering
  \caption {MPFP-NET AND THE OTHER MODELS PERFORMANCE ON THE LEVIR DATASET \cite{zou2017random}. PARAMS. AND FLOPS (IN BILLIONS) REPRESENT THE NUMBER OF THE PARAMETERS AND MULTIPLY-ADDS. LAT SIGNIFIES INFERENCE LATENCY WITH BATCH SIZE 1. THE MODELS THAT HAVE ALMOST EQUAL PERFORMANCE ARE GROUPED.}
  \label{tab:3}
  \begin{tabular}{ll | ll | ll | ll | llll}     \hline
 Model & mAP & Params & Ratio & FLOPs & Ratio & GPU LAT(ms) & Speedup & CPU LAT(s) & Speedup \\    \hline

{\bf MPFP-Net-S$_0$ (512)} & {\bf 62.39} & {\bf 4.1M} & {\bf 1$\times$} & {\bf 3.8B} & {\bf 1$\times$} & {\bf 19$\pm$2.1} & {\bf 1$\times$}  & {\bf 0.35$\pm$0.003} & {\bf 1$\times$}  \\ 
LARGE-RAM \cite{zou2017random} & 62.42 & 18.3M & 4.3$\times$ & 53B & 13.6$\times$ & 52 & 2.7$\times$ & 3.9$\pm$0.034 & 9.6$\times$    \\ \hline 

{\bf MPFP-Net-S$_1$ (640)} & {\bf 65.07} & {\bf 7.6M} & {\bf 1$\times$} & {\bf 8.1B} & {\bf 1$\times$} & {\bf 23$\pm$0.8} & {\bf 1$\times$} & {\bf 0.78$\pm$0.004} & {\bf 1$\times$}   \\
RIRBM \cite{diao2016efficient} & 64.31	& 67M & 8.5$\times$ & 368B & 46.1$\times$ & 279$\pm$0.3 & 12.1$\times$ & 57$\pm$0.017 & 63.1$\times$   \\ \hline

{\bf MPFP-Net-S$_2$ (768)} & {\bf 72.55} & {\bf 8.7M} & {\bf 1$\times$}  & {\bf 16.3B} & {\bf 1$\times$}  & {\bf 26$\pm$1.1} & {\bf 1$\times$} & {\bf 1.3$\pm$0.003} & {\bf 1$\times$} \\
SUCNN \cite{hu2019sample} &  70.11 & 103M & 11.4$\times$ & 1028B & 64.3$\times$ & 362$\pm$0.4 & 13.8$\times$ & 57$\pm$0.028 & 35.2$\times$   \\  \hline

{\bf MPFP-Net-S$_3$ (896)} & {\bf 75.71} & {\bf 13.4M} & {\bf 1$\times$} & {\bf 35B} & {\bf 1$\times$} & {\bf 44.8$\pm$0.5} & {\bf 1$\times$} & {\bf 2.7$\pm$0.005} & {\bf 1$\times$} \\
HSF-Net \cite{li2018hsf} &  75.33 &	152M&	11.3$\times$ &	2389B&	68.4$\times$ &	566$\pm$0.8&	12.5$\times$ &	98$\pm$0.043&	32$\times$ \\
ResNet-50 + NAS-FPN (1280) \cite{ghiasi2019fpn} &  73.75&	59.6M&	4.5$\times$&	350B&	10$\times$&	267$\pm$0.4&5.8$\times$&57$\pm$0.170&20.2$\times$ \\
ResNet-50 (1280) \cite{he2016deep} &  --&	25.6M&--	&125B&	-- &	117$\pm$0.3&-- &24$\pm$0.05&-- \\ \hline

{\bf MPFP-Net-S$_4$ (1024)} & {\bf 79.05} &	{\bf 21.1M} & {\bf 1$\times$} &	{\bf 56B}&	{\bf 1$\times$} & {\bf 76$\pm$0.4}& {\bf 1$\times$} & {\bf 5.1$\pm$0.002}& {\bf 1$\times$} \\
Sig-NMS \cite{dong2019sig} &  78.11&	71.7M&	3.8$\times$ &	479B&	8.3$\times$ &	290$\pm$1.3&	3.8$\times$ &	60.14$\pm$0.007&	11.9$\times$   \\
ResNet-50 + NAS-FPN (1280$@$84) \cite{ghiasi2019fpn} &  74.28& 72.6M & 3.6$\times$ &	546B&	9.5$\times$ &327$\pm$0.1&4.3$\times$ &63.71$\pm$0.031&12.5$\times$  \\ \hline

{\bf MPFP-Net-S$_5$ (1280)} &  {\bf 83.42}&	{\bf 34.2M}& {\bf 1$\times$}& {\bf 142B}& {\bf 1$\times$}& {\bf 145$\pm$2.1}& {\bf 1$\times$}&	{\bf 11.4$\pm$0.014}& {\bf 1$\times$}   \\
LV-Net \cite{li2019nested} &    82.34	& 114M &	3.3$\times$&	1259B&	8.8$\times$&	438$\pm$1.6&	3.2$\times$&	84.31$\pm$0.034&	7.5$\times$   \\ \hline

{\bf MPFP-Net-S$_6$ (1280)} & {\bf 86.73} &	{\bf 51.2M}& {\bf 1$\times$}&	{\bf 228B}&	{\bf 1$\times$}& {\bf 198$\pm$0.7}& {\bf 1$\times$}& {\bf 17$\pm$0.002}& {\bf 1$\times$} \\
HSP \cite{xu2020hierarchical} &    85.32&	125M &	2.6$\times$&	1428B&	6.2$\times$&416$\pm$1.8&2.25$\times$&85$\pm$0.096&	4.4$\times$  \\
MS-OPN \cite{deng2018multi} &    85.21&	101.6M &	2.2$\times$&	1019B&	4.4$\times$&380$\pm$1.5&1.9$\times$&72$\pm$0.081&	4.2$\times$  \\  \hline
\end{tabular}
\end{table*}
\subsection{Dataset}
\subsubsection{NWPU VHR-10} For evaluating the performance of the proposed MPFP-Net model, we use the challenging ten class Northwestern Polytechnical University Very High-Spatial Resolution (NWPU VHR-10) dataset \cite{dong2019sig}. This dataset contains 650 VHR optical RSIs, in which 565 images were obtained from Google Earth, where each image has the size of $1000\times 1000$ pixels with the resolution ranging from $0.5\sim2$ m, and 85 pansharpened infrared images with $0.08 m$ resolution. The dataset includes ten manually annotated classes. 
\subsubsection{LEVIR} This dataset contains 22k high resolution Google Earth images, where each image has the size of $800\times 600$ pixels and a resolution of $0.2\sim1.0$ m/pixel \cite{zou2017random}. The dataset contains three annotated classes with small objects of $30\times 30$ pixels and minimum objects of $10\times 10$ pixels.
\subsubsection{DOTA} It is a large RSIs dataset \cite{xia2018dota} used for objects detection which comprises of 2806 images with different size ranges ($800\times 800$ to $4000\times 4000$ pixels) and 188282 instances of 15 categories of objects: plane, baseball diamond (BD), bridge, ground field track (GFT), harbor and helicopter (HC), small vehicle (SV), large vehicle (LV), tennis court (TC), basketball court (BC), storage tank (ST), soccer ball field (SBF), roundabout (RA), and swimming pool (SP). Each image is labeled with an arbitrary quadrilateral. 

\subsection{Evaluation Metrics}
The target detection results contain two components, bounding boxes (BBs) which enclose the detected targets and the labels. In general, IoU is used as the evaluation metrics in object detection which denotes the ratio between the estimated object and the ground truth BBs. The IoU is formulated as follows:
\begin{eqnarray}
IoU=\frac{(S_{BBs}\cap S_{GT})}{(S_{BBs}\cup S_{GT})}      
\label{eq:23}
\end{eqnarray}
in which $S_{BB}$ and $S_{GT}$ are the predicted areas and the ground truth boxes, respectively. Here, we use precision recall curves and the average value of precision (AP) for a single object class from recall 0 to 1 to assess the object detection framework, in which a higher AP represents better performance. While having multiclass objects, mAP is used to calculate the average AP of all the classes.
\begin{figure*}
  \centering
  \includegraphics[width=6.8in, height=2.8in]{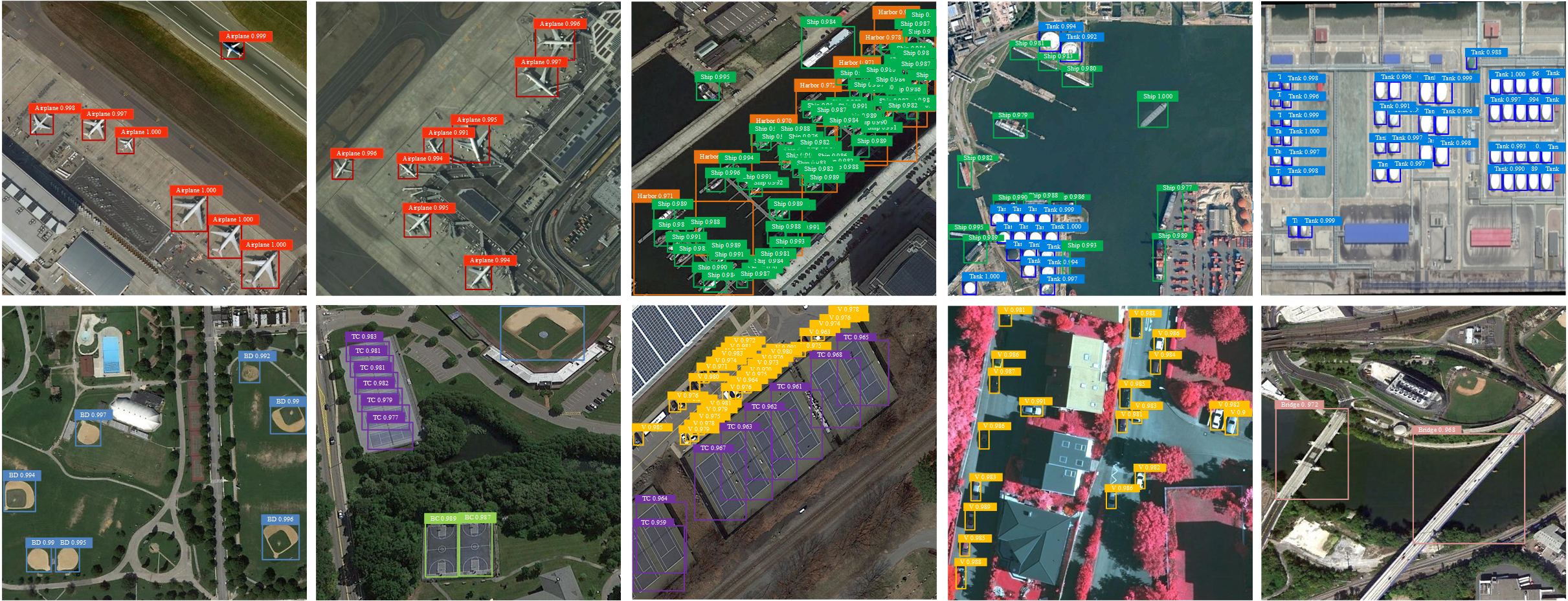}
\caption{Selected detection samples by MPFP-Net on NWPU VHR-10.(zoom in for better vision)} 
\label{fig:8}
\end{figure*}
\subsection{Training procedures, and implementation details}
NWPU-VHR-10 and LEVIR labels are in a conventional axis-aligned BBs form, while DOTA objects' labels are in a quadrilateral form. For adapting the both settings, our proposed MPFP-Net uses both horizontal and oriented BBs (HBB, OBB) as ground truth, where HBB:$\{x_{min},y_{min},x_{max},y_{max}\}$, OBB:$\{x_{center},y_{center},w,h,\theta \}$, here $w$, $h$ denote width, height and $\theta$ is within $[0, 90^{\circ})$ to create ground truth for each object. In training, the OBB ground truth is produced by a group of rotated rectangles which properly overlap with the given quadrilateral labels.
For the NWPU-VHR-10 and LEVIR datasets, the MPFP-Net just produces HBB results, as OBB ground truth does not exist in the datasets. 

\begin{figure*}
\centering
\includegraphics[width=6.8in, height=2.8in]{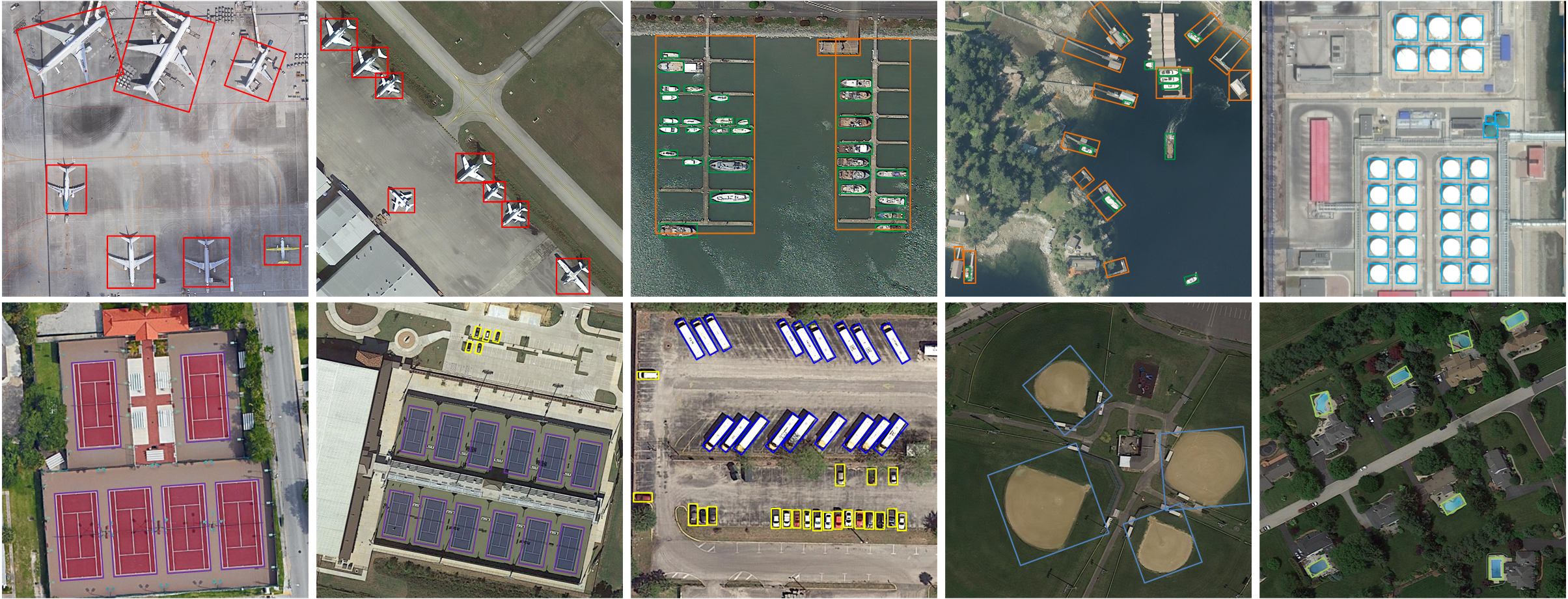}
\caption{Example of detection results by the MPFP-Net On DOTA.}
\label{fig:15}
\end{figure*}

However, for the DOTA, the MPFP-Net produces both HBB and OBB outputs, as is presented in Fig. \ref{fig:15}.
In the training phase, for the LEVIR and NWPU VHR-10 datasets, we resize the original images to $512\times 512$ pixels. For the NWPU VHR-10, the quantity of images is insufficient, to increase the training set, we perform rotation, and mirroring. For the DOTA dataset, we split the images into $640\times 640$ patches with 200 pixels overlap by the development toolkit. In our experiments, 75\% of LEVIR is selected for training and the remaining 25\% for testing. Also, we select 60\% of NWPU-VHR for training, 10\% for validation, and the rest for testing, and 60\% of the DOTA dataset for training, 20\% for validation, and the remaining for testing. We employ the ResNet-50 + SPN \cite{shamsolmoali2008road} as backbone. For the DOTA and LEVIR datasets, we train the model for 150k iterations with batch size 1 on 2 Tesla P40 GPUs, which took around 24 hours. The initial learning rate is set to 1e-4 and is divided by 10 after every 30k iterations. The weight decay is set to 0.0005 and the momentum is 0.9, batch normalization and Swish activation (to enhance the backpropagation) are used after each convolution layer. Adam optimization \cite{kingma2014adam} is used to speed up the training. On the other hand, for the NWPU dataset, we train the model with 30k iterations, and the initial learning rate was set to 1e-3 and changed to 1e-4 and 1e-5 at 10k and 20k iterations, respectively, which took around 3 hours. For the NWPU dataset since the number of images is not enough, during training, we also use data augmentation including rotation and random flip.
\begin{figure*}
\centering
\includegraphics[width=6.7in]{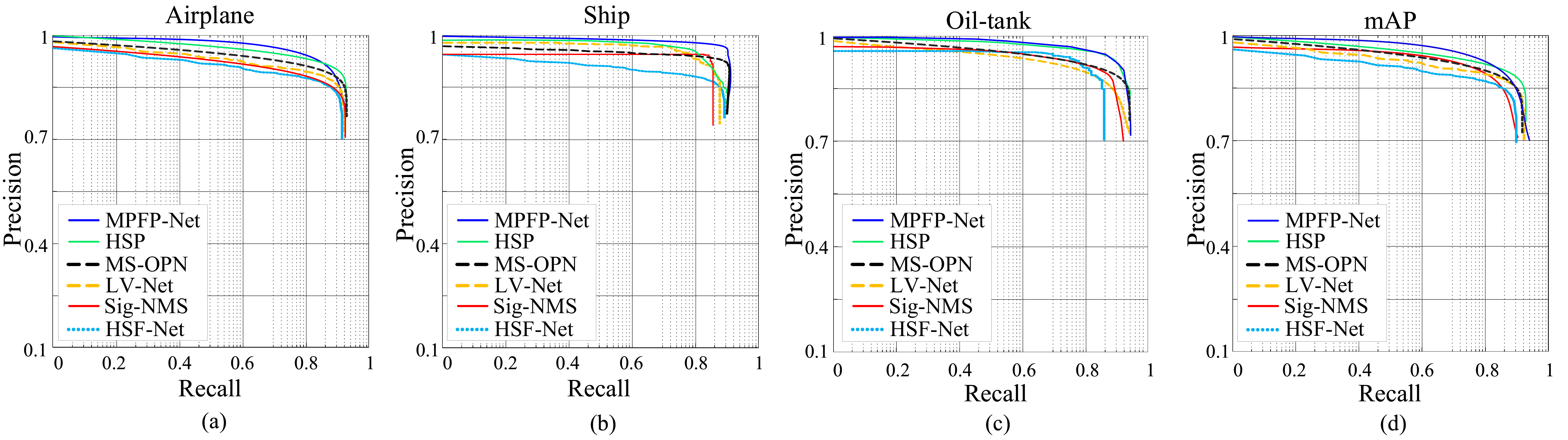}
\caption{Precision–Recall comparisons between MPFP-Net and other methods on LEVIR dataset, for the classes of airplane,ship, oil-tank, and Mean AP.}
\label{fig:9}
\end{figure*}

In Table \ref{tab:3}, we compare MPFP-Net with several object detection methods on LEVIR. Our proposed MPFP-Net achieves better accuracy with lower computation costs, compared to the other models across a broad range of resource constraints. The MPFP-Net-S$_0$ generates only 4.1M parameters on top of the backbone parameters and obtains almost the same accuracy as LARGE-RAM \cite{zou2017random} with $13.6\times$ fewer FLOPs. In comparison with RIRBM \cite{diao2016efficient}, the proposed MPFP-Net-S$_1$ obtains a better detection rate with up to $46\times$ lower FLOPs and $8.5\times$ less number of parameters. In our proposed model, by increasing the input size from $512 \times 512$ to $640 \times 640$ the result is improved with 3 mAP, on the other hand, the model’s size and Flops are increased by 1.8\% and 2.1\% respectively.
\begin{table} 
\centering
  \caption{PERFORMANCE COMPARISON FOR HBB TASK ON THE NWPU-VHR-10. THE BEST RESULTS ARE BOLD.}
\label{tab:4}
  \begin{tabular}{p{15mm}p{8mm}p{7mm}p{7mm}p{6mm}p{6mm}p{7mm}}     \hline
   & HSF-Net & Sig-NMS & LV-Net & MS-OPN & HSP & MPFP-Net \\   \hline
Airplane&	88.64&	90.84&	93.40&	99.74& {{99.80}}&\textbf{99.84} \\[-0.2em]
Ship&		80.54	&80.58&	81.80& 89.02 & {92.47}&	\textbf{92.63}  \\[-0.2em]
ST&	59.13	&59.25 &	73.57 & 80.13& \textbf{96.99}& {96.98} \\[-0.2em]
BD&	88.35&{90.89}&	98.35&	98.10& \textbf{98.58}&{98.49}  \\[-0.2em]
TC&	79.38&	80.86 &	84.89 & 86.09& \textbf{90.39}&	{89.83}   \\[-0.2em]
BC&	89.87&	90.94&	88.86& \textbf{92.57}& {91.49}&{91.96}  \\[-0.2em]
GTF&	96.41&	\textbf{99.85}&	96.87& 98.65& {99.08}&{99.73}  \\[-0.2em]
Harbor&	81.81&	90.39&	91.64&{94.61}& 88.93&\textbf{94.82}   \\[-0.2em]
Bridge&	65.59&	67.86&	82.91& \textbf{94.37}& 87.11&{92.30}  \\[-0.2em]
Vehicle&	78.89&	78.16&	79.87& 82.23& {89.09}&\textbf{89.15}  \\[-0.2em]
mAP&	80.84&	82.93&	87.19& 91.53& {93.40}& \textbf{94.57}   \\[-0.2em]
\hline
Parameters&	153M&{71.7M}&	115M&	103M&126M &\textbf{52.3M}  \\[-0.2em]
FLOPs	&	2389B&{479B}&	1270B&	1022B&1525B &\textbf{230B}   \\[-0.2em]
\hline
\end{tabular}
\end{table}
\subsection{Comparison with other State-of-the-Art Methods}
Fig. \ref{fig:8} presents some detected objects by our proposed model on the evaluation datasets.
\subsubsection{Performance evaluation on the NWPU VHR dataset} As clearly seen in Fig. \ref{fig:8}, the detected objects are belonging to various classes, for example, airplanes, ships, harbors, and oil tanks. The proposed MPFP-Net has superior performance in object detection and even the objects that stay very close together are also properly detected. 
\begin{table} 
\centering
\caption{PERFORMANCE COMPARISON FOR HBB TASK ON THE LEVIR DATASET.}
\label{tab:5}
\begin{tabular}{p{10mm}p{8mm}p{7mm}p{7mm}p{6mm}p{6mm}p{7mm}}     \hline
& HSF-Net & Sig-NMS & LV-Net & MS-OPN & HSP& MPFP-Net \\   \hline
Airplane &	81.04 &86.83 &	82.58 &	85.09 & {86.92} &\textbf{87.24}   \\[-0.2em]
Ship &	77.29 &	79.42 &	80.57 &{83.76} &83.75 &\textbf{85.57} \\[-0.2em]
Oil-tank &	67.62 &	68.28	 &83.59 &86.80 & \textbf{87.42}&{87.35} \\[-0.2em]
mAP &	75.33 &	78.11	 &82.34 &85.21 & {85.32}&\textbf{86.73} \\[-0.2em]
\hline
Parameters &	152M &{71.7M} &114M &	101.6M &125M&\textbf{51.2M} \\[-0.2em]
FLOPs	 &	2389B	 &{479B} & 	1259B &	1019B	 &1428B &\textbf{228B} \\[-0.2em]
\hline
\end{tabular}
\end{table}
\begin{table}[!t]
\centering
\vspace{-3mm}
\caption{PERFORMANCE COMPARISON FOR HBB TASK ON THE DOTA DATASET.}
  \label{tab:6}
 \begin{tabular}{p{10mm}p{8mm}p{7mm}p{7mm}p{6mm}p{6mm}p{7mm}}     \hline
  & HSF-Net & Sig-NMS & LV-Net & MS-OPN& HSP & MPFP-Net \\    \hline
Plane &	82.34&	83.67&	85.38& 88.61& {90.36}& \textbf{90.49}  \\[-0.2em]
BD&	79.66&	80.29&	84.63& 86.11& \textbf{86.85}& {86.79} \\[-0.2em]
Bridge&	48.74	&50.12&52.81	&54.76&{62.51}&\textbf{62.68} \\[-0.2em]
GTF&	76.38	&77.51&	79.49& \textbf{79.96}& {79.83}& 79.71  \\[-0.2em]
SV&	65.42	&68.37&	71.64&75.29&{78.07}&\textbf{78.22}  \\[-0.2em]
LV&	64.92	&69.74&	70.83&{78.34}& 81.79&\textbf{81.97}  \\[-0.2em]
Ship&	75.37	&77.46&	80.72&83.17& \textbf{85.27}&{85.24}  \\[-0.2em]
TC&	84.68	&86.69&	88.48&\textbf{90.91}& 90.82&{90.88}  \\[-0.2em]
BC&	79.42	&81.45&	84.39&{86.12}& \textbf{87.24}& 87.21  \\[-0.2em]
ST&	74.19	&79.76&	81.73&84.96& {85.90}&\textbf{85.98}  \\[-0.2em]
SBF&	60.33&61.94&	65.75&68.58& {69.93}&\textbf{70.18}  \\[-0.2em]
RA&	65.53&68.44&	70.51&71.27& \textbf{72.08}&{72.02}  \\[-0.2em]
Harbor&	69.58	&71.45&	73.40&76.12& {84.11}&\textbf{84.24}  \\[-0.2em]
SP&	68.92	&68.64&	70.85&71.59& {80.92}&\textbf{81.13}  \\[-0.2em]
HC&	64.60	&66.39&	66.42&67.77& \textbf{69.81}&{69.78}  \\[-0.2em]
mAP& 70.67&	72.65&	75.23	&77.38&{80.36}&\textbf{80.43}  \\[-0.2em]
\hline
Parameters&	155M&\text{72.8M}	&116M&	105M	&128M&\textbf{54.8M} \\[-0.2em]
FLOPs&	2394B&\text{481B}&	1286B	&1028B&1529B&\textbf{242B} \\[-0.2em]
\hline
\end{tabular}
\end{table}
For the evaluation, other state-of-the-art RS detection approaches such as Sig-NMS \cite{dong2019sig}, HSF-Net \cite{li2018hsf}, LV-Net \cite{li2019nested}, MS-OPN \cite{deng2018multi}, and HSP \cite{xu2020hierarchical}, are applied to the NWPU VHR-10 and their object detection performance are compared with that of MPFP-Net. Table \ref{tab:4} lists the HBB detection results. 
As the results show, the MPFP-Net has superior performance in comparison with the other methods, and achieves better results in the airplane and ship classes. The airplane and ship classes contain tiny targets, indicating that our proposed method performs satisfactorily in detecting small objects. Our proposed MPFP-Net retains better mAP with much fewer parameters and FLOPs as compared to the other approaches on the NWPU-VHR-10 dataset.
\subsubsection{Performance evaluation on LEVIR} detailed object detection comparisons between MPFP-Net and the other state-of-the-art models on the LEVIR dataset are reported in Table \ref{tab:5}. It is clear that MPFP-Net shows the best detection results in all the classes. In particular, our model surpasses HSP and MS-OPN in airplane and ship classes, which have a large number of small objects, and this indicates that MPFP-Net is able to successfully detect small objects. Fig. \ref{fig:9} shows the PRCs of the three classes of LEVIR, and mean AP by HSF-Net, Sig-NMS, LV-Net, MS-OPN, HSP and MPFP-Net, respectively. The blue curves which represent the performance of our proposed model are higher or similar to those of different classes.
\subsubsection{Performance evaluation on DOTA} In Table \ref{tab:6}, we evaluate the HBB detection performance of MPFP-Net on 15 classes of the DOTA dataset in comparison with the other approaches.The results demonstrate that the proposed method achieves state-of-the-art results and comparable performance with the other object detection approaches. 

\begin{figure}
\centering
\includegraphics[width=3.5in]{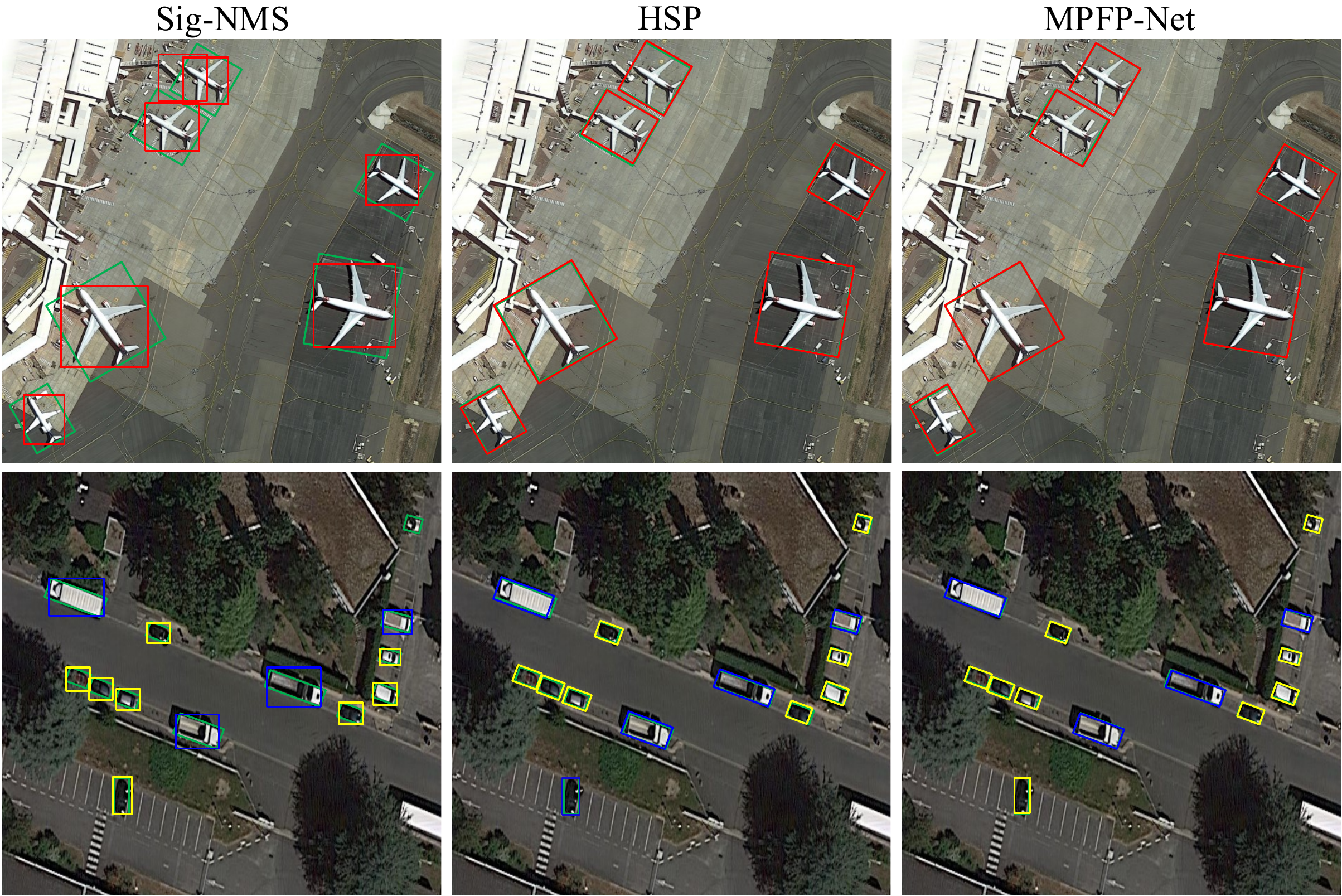}
\caption{Qualitative detection comparison by different models on the DOTA dataset. The green boxes, show ground truth. The red, blue and yellow boxes indicate the detected planes, large vehicles and small vehicles respectively.}
\label{fig:16}
\end{figure}

To show the advantages of MPFP-Net over the other methods, we perform a qualitative comparison of different methods on the DOTA dataset. The results are shown in Fig. \ref{fig:16}. 
The results show the other approaches cannot robustly detect the objects in images, and the background is mis-detected as the foreground. Moreover, in the other methods, the bounding boxes are not well fit to the detected objects, whereas Sig-NMS \cite{dong2019sig} only generates the horizontal box. However, our method can stably produce precise results.

%
\subsection{Experiment on Large-Scale Images}
\begin{figure}
\centering
\includegraphics[width=3.5in]{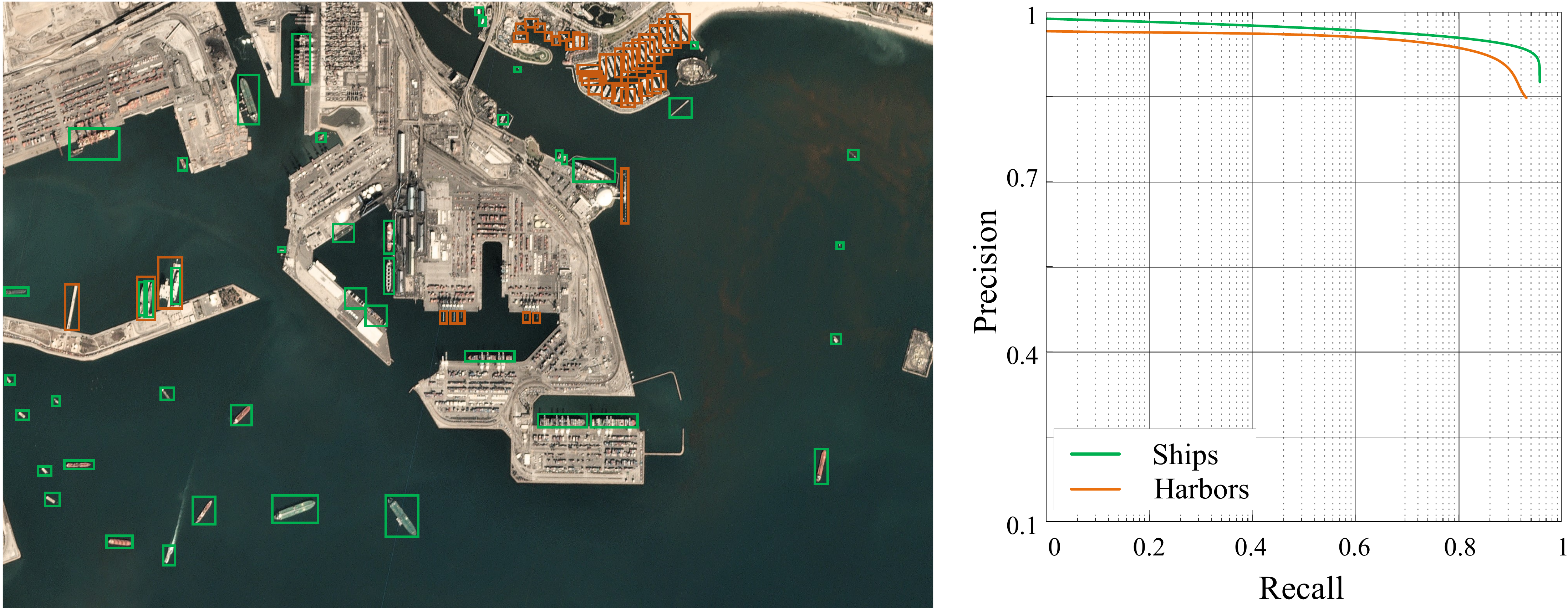}
\caption{Detection results on a large-scale RSI from Geomatica-CGI System. Green and orange respectively show ships and harbors.}
\label{fig:10}
\end{figure}

In Fig. \ref{fig:10}, we test MPFP-Net performance on a large-scale RSI. It is observed that the pre-trained MPFP-Net has an adequate flexibility on different image sources and conditions, which shows the effect of multiple patches and feature pyramid learning.
\begin{table}[t]
\centering
  \caption{PROPOSED MODEL EVALUATION WITH DIFFERENT BACKBONES.}
  \label{tab:7}
 \begin{tabular}{p{35mm}p{9mm}p{12mm}p{8mm}}     \hline
 Configuration & mAP& Parameters & Flops  \\   \hline
MPFP-Net-S$_6$+ FPN&	78.41&	122M&	1253B  \\
MPFP-Net-S$_6$+ SPN&	83.69&	84.9M	&387B  \\
MPFP-Net-S$_6$+ SPN+ ESS&	86.73	&51.2M	&228B  \\
\hline
\end{tabular}
\end{table}
\begin{figure*}
  \centering
  \includegraphics[width=6.4in]{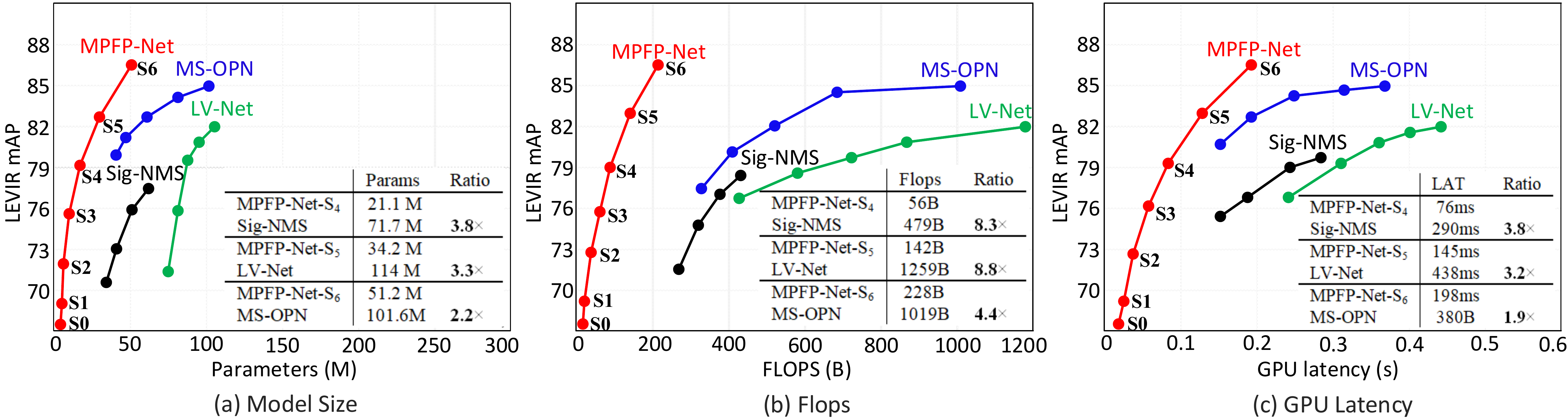}
\caption{Comparison with different model size and inference latency. Latency is measured with batch size 1 on the machine equipped with a P40 GPU. The proposed MPFP-Net models are 2.2$\times$ $\sim$ 3.8$\times$ smaller, and 1.9$\times$ $\sim$ 3.7$\times$ faster than the other detectors.}
\label{fig:11}
\end{figure*}
\begin{figure}
  \includegraphics[width=3.4in]{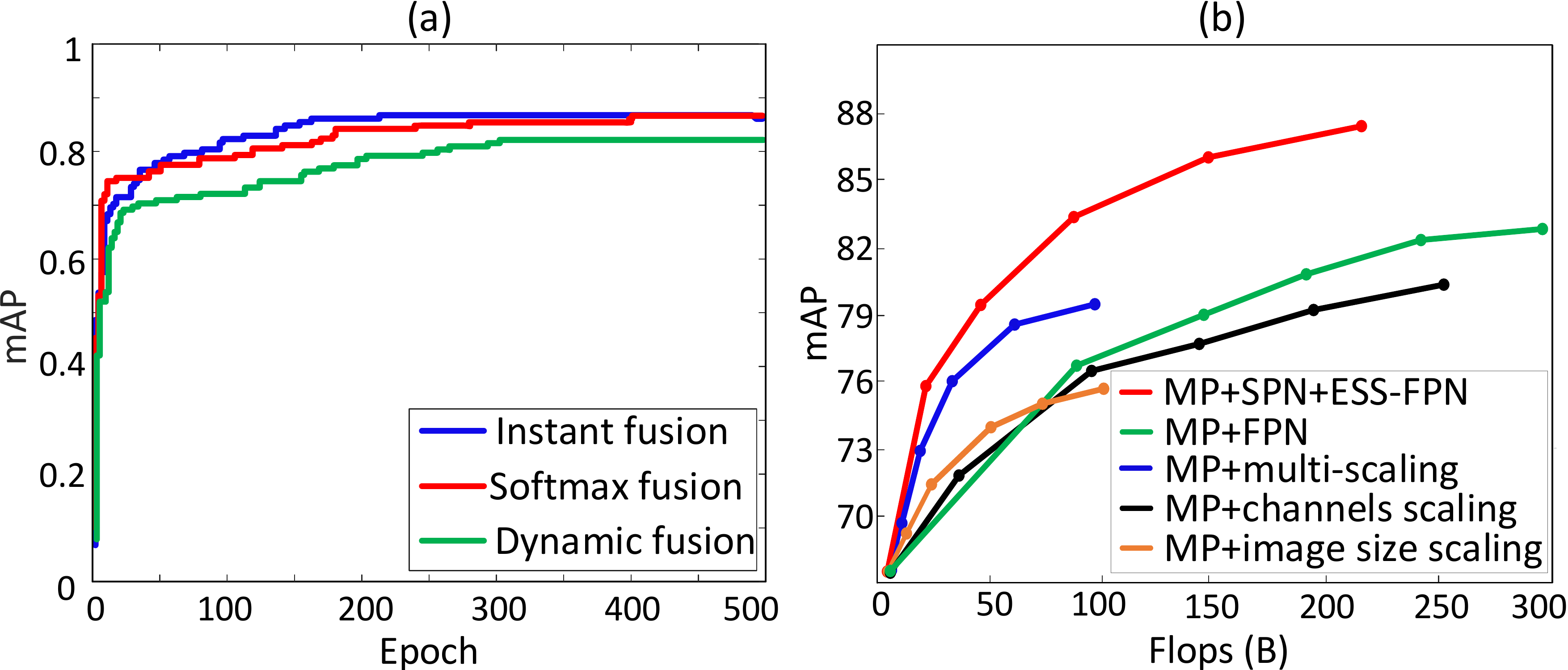}
\caption{(a) Instant fusion vs softmax fusion and dynamic fusion. (b) Comparison of different Multi-patch scaling methods.}
\label{fig:13}
\end{figure}
\section{ABLATION STUDY}
\label{sec:5}
This section ablates different components of MPFP-Net on the LEVIR test set. We evaluate the case of different parameter sizes, FLOPs and latency on P40 GPU. Each model is run 5 times with the batch size of 1 and the mean and standard deviation are reported in Table \ref{tab:3}. Fig. \ref{fig:11} shows the model's size, Flops and GPU latency. In comparison with the other models, MPFP-Net is up to 3.7$\times$ faster. We conduct a set of evaluations to measure the contribution of the backbone network and ESS-FPN to the detection accuracy and efficiency improvements of the proposed MPFP-Net. Table \ref{tab:7} reports the impact of each module on the overall performance of our proposed model. 
Starting from MPFP-Net-S$_6$ with FPN \cite{lin2017feature} as the backbone, firstly, instead of FPN we adopt the SPN \cite{shamsolmoali2008road}, which improves the detection accuracy by more than 5 mAP with a smaller number of parameters and FLOPs. Later, we add the proposed ESS-FPN, where the detection performance is improved by 3 mAP with less parameters and FLOPs.
\subsection{Instant fusion vs Softmax and Dynamic fusion}
As earlier discussed, we propose an instant feature fusion approach that preserves the benefits of the normalized weights while reducing the softmax computation cost. In Fig. \ref{fig:13}(a), we examine the performance of MPFP-Net while adopting dynamic fusion, softmax fusion and the proposed instant fusion. The proposed instant fusion achieves accuracy and learning behavior similar to the softmax fusion, however runs $1.24\times$ faster. During the training, the normalized weights quickly change, suggesting that various features unequally contribute to the feature fusion. 
\subsection{Multi-Patch and Combined Multi-Scaling}
As discussed in Sections \ref{sub:2}, we propose multiple-patch learning to deal with the lack of full object instance-labelling and propose a combined multiple scaling method to increase all the dimensions of ESS-FPN for better learning and consequently increasing the detection performance of our proposed model. In Fig. \ref{fig:13}(b), we compare our approach with the other models that only use patch learning or scale up a single dimension (resolution/depth/width). As the results illustrate, our model results in better efficiency as compared to the other baselines, which signifies the advantages of patch-wise and jointly scale-wise learning. 
\section{CONCLUSION}
\label{sec:6}
In this paper, we have proposed a novel weakly supervised model for detecting multi-scale objects in RSIs using a multi-patch feature pyramid network. First, we integrated automatic patch selection, feature aggregation and semantic domain projection within a single unified framework. Second, we proposed a weighted feature pyramid network which uses multi-directional connections for fast and efficient scale wise feature fusion to further optimize object detection in RSIs. Moreover, a joint loss was used to train the whole network end-to-end. On the basis of these methodologies, we introduced a new detector, named MPFP-Net, which obtains better accuracy and efficiency than the other state-of-the-art methods on RSIs. To evaluate the performance of MPFP-Net for multi-scale objet detection, three publicly available datasets were used. On these datasets, we evaluated the performance of our proposed method, compared with several CNNs based object detection models. Experimental results demonstrate that MPFP-Net can effectively and efficiently detect multi-scale objects.

\bibliographystyle{IEEEtran}
\bibliography{IEEEabrv,IEEEexample}
\end{document}